\documentclass[conference]{IEEEtran}
\IEEEoverridecommandlockouts

\usepackage{cite}
\usepackage{amsmath,amssymb,amsfonts}
\usepackage{algorithmic}
\usepackage{graphicx}
\usepackage{textcomp}
\usepackage{xcolor}
\def\BibTeX{{\rm B\kern-.05em{\sc i\kern-.025em b}\kern-.08em
    T\kern-.1667em\lower.7ex\hbox{E}\kern-.125emX}}

\usepackage{hyperref}

\usepackage{algorithmic}
\usepackage[ruled, vlined, linesnumbered, commentsnumbered]{algorithm2e}
\usepackage{amsmath}
\usepackage{amsfonts}
\usepackage{amsthm}
\usepackage{amstext}
\usepackage{color}
\usepackage[capitalise]{cleveref}
\usepackage{booktabs}
\usepackage{enumitem}
\usepackage{graphicx}
\usepackage{multirow}
\usepackage{subcaption}
\usepackage{tabularx}
\usepackage{adjustbox}
\usepackage{pgfplots}
\pgfplotsset{compat=newest}

\usepackage{amsmath}
\usepackage{graphicx}
\usepackage[utf8]{inputenc}
\usepackage[english]{babel}

\newcommand{\UserSpace}{\cU}
\newcommand{\ItemSpace}{\cI}

\newcommand{\ClickRv}{c}

\newcommand{\ConvRv}{v}

\newcommand{\ClickMat}{C}

\newcommand{\ConvMat}{V}

\newcommand{\ExposeSpace}{\cD}
\newcommand{\ClickSpace}{\cC}
\newcommand{\NonClickSpace}{\cN}
\newcommand{\ConvSpace}{\cV}

\newcommand{\CounterNonClickSpace}{\cN^{*}}

\newcommand{\FeatRv}{x}
\newcommand{\FeatMat}{X}

\newcommand{\ours}{\text{ESCIM}}

\newcommand{\bbE}{\mathbb{E}}

\newcommand{\bbP}{\mathbb{P}}

\newcommand{\cC}{\mathcal{C}}
\newcommand{\cD}{\mathcal{D}}

\newcommand{\cI}{\mathcal{I}}

\newcommand{\cL}{\mathcal{L}}

\newcommand{\cN}{\mathcal{N}}

\newcommand{\cS}{\mathcal{S}}

\newcommand{\cU}{\mathcal{U}}
\newcommand{\cV}{\mathcal{V}}

\usepackage{tikz}
\usepackage{pgf}
\usetikzlibrary{positioning,shapes,shapes.misc,arrows,calc,arrows.meta,fit, backgrounds,quotes,intersections,patterns,decorations.pathmorphing,decorations.pathreplacing,decorations.markings,tikzmark}
\tikzset{
  >={Latex[width=1.6mm,length=2mm]},
  font=\sffamily\footnotesize,
  RR/.style={draw,circle,inner sep=0mm, minimum size=5.mm,font=\sffamily\footnotesize}
}

\begin{document}

\title{On Predicting Post-Click Conversion Rate via Counterfactual Inference}

\author{\IEEEauthorblockN{Junhyung Ahn$^*$\thanks{$^*$ Work done at Dable.}}
\IEEEauthorblockA{\textit{Naver Corp.} \\
Seongnam, South Korea\\
junhyung.ahn@navercorp.com}
\and
\IEEEauthorblockN{Sanghack Lee$^\dagger$ \thanks{$^\dagger$ Corresponding author}}
\IEEEauthorblockA{\textit{Seoul National University} \\
Seoul, South Korea\\
sanghack@snu.ac.kr}
}

\maketitle

\begin{abstract}
\label{sec:abs}
Accurately predicting conversion rate (CVR) is essential in various recommendation domains, such as online advertising systems and e-commerce.
These systems utilize user interaction logs, which consist of exposures, clicks, and conversions.
CVR prediction models are typically trained solely based on clicked samples, as conversions can only be determined following clicks.
However, the sparsity of clicked instances necessitates the collection of a substantial amount of logs for effective model training.
Recent works address this issue by devising frameworks that leverage non-clicked samples.
While these frameworks aim to reduce biases caused by the discrepancy between clicked and non-clicked samples, they often rely on heuristics.
Against this background, we propose a method to counterfactually generate conversion labels for non-clicked samples by using causality as a guiding principle, attempting to answer the question, ``Would the user have converted if he or she had clicked the recommended item?".
Our approach is named the \textbf{\underline{E}}ntire \textbf{\underline{S}}pace \textbf{\underline{C}}ounterfactual \textbf{\underline{I}}nference \textbf{\underline{M}}ulti-task Model (ESCIM).
We initially train a structural causal model (SCM) of user sequential behaviors and conduct a hypothetical intervention (i.e., click) on non-clicked items to infer counterfactual CVRs.
We then introduce several approaches to transform predicted counterfactual CVRs into binary counterfactual conversion labels for the non-clicked samples.
Finally, the generated samples are incorporated into the training process.
Extensive experiments on public datasets illustrate the superiority of the proposed algorithm.
Online A/B testing further empirically validates the effectiveness of our proposed algorithm in real-world scenarios.
In addition, we demonstrate the improved performance of the proposed method on latent conversion data, showcasing its robustness and superior generalization capabilities.
The code for the proposed framework is available at \href{https://github.com/JunhyungAhn/ESCIM}{https://github.com/JunhyungAhn/ESCIM}.
\end{abstract}

\begin{IEEEkeywords}
CVR Prediction, Structural Causal Model, Counterfactual Inference
\end{IEEEkeywords}

\section{Introduction}
\label{sec:intro}
\begin{figure}[t]
    \centering
    \includegraphics[width=\columnwidth]{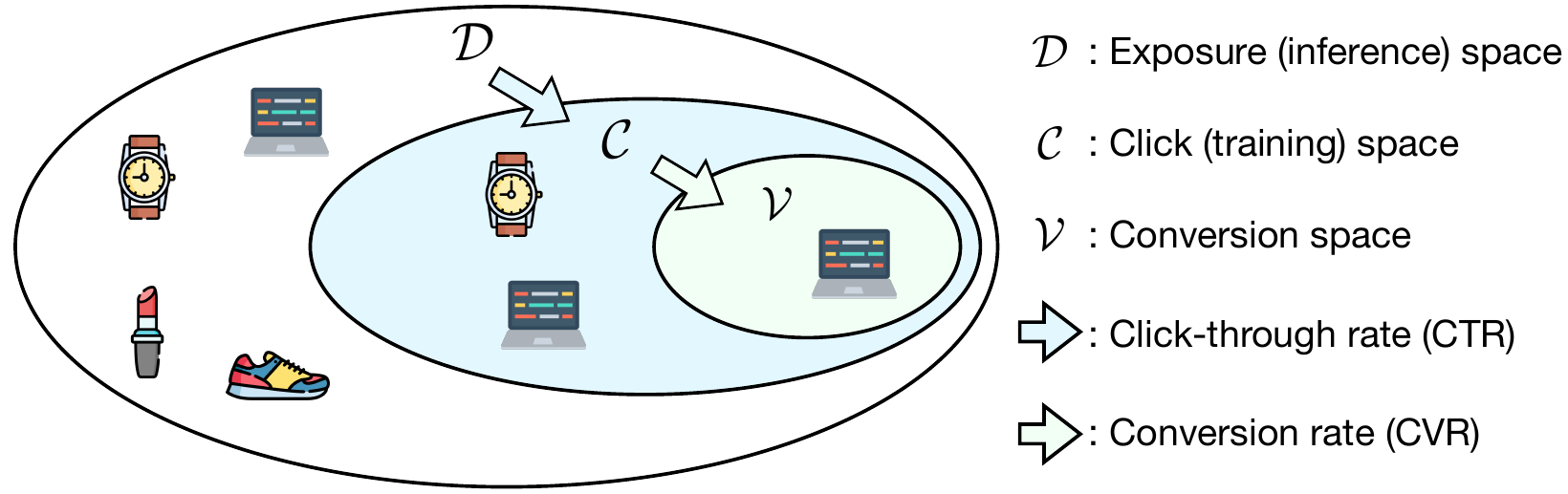}
    \caption{An example of the data sparsity and selection bias of the CVR estimation task, where the training space $\ClickSpace$ only contains clicked samples, while the inference space $\ExposeSpace$ consists of all exposed samples.}
    \label{fig:ssb_ds}
\end{figure}

Recommender systems have emerged as powerful tools in various industrial domains, such as social networks~\cite{Zhou19Dien}, online e-commerce~\cite{Ma18Esmm, Wen20Esm2}, and advertising~\cite{Pan18Fwfm}.
The objective of recommender systems is to suggest the most preferred items to users to induce clicks or conversions (e.g., purchases) from them.
To achieve this objective, recommender models are trained with logs of numerous user and item pairs to characterize the underlying relationships in users' click and purchase behavior.
A typical sequence of user behavior in online e-commerce follows the path: ``exposure $\rightarrow$ click $\rightarrow$ conversion"~\cite{Ma18Esmm}. 
For the space of items, this can be illustrated as a Venn diagram (\cref{fig:ssb_ds}).
Metrics that capture user behavior include click-through rate (CTR), post-click conversion rate (CVR), and click \& conversion rate (CTCVR).
Given that CVR is defined solely within the click space, a naive approach involves training the CVR estimator using only the clicked samples. 
This approach faces two critical challenges: sample selection bias and data sparsity.

In recommendation system literature, sample selection bias refers to the distortion that occurs when the data used for training the model is missing not at random (MNAR).\footnote{We formally adopt counterfactual inference, which is a problem in the field of causal inference. In the field, MNAR deals with cases where some values are missing in the data, which does not match the usage of the term MNAR in the recommender system. Yet, we will follow the convention in the recommender system.} 
This means that the observed interactions, such as clicks or purchases, depend on unobserved factors, leading to an overrepresentation of popular items and active users. 
In contrast, items with lower CVR are less likely to be clicked, making them less likely to be included in the training dataset. 
Consequently, the model may struggle to predict interactions for less active users or niche items accurately, resulting in suboptimal recommendations.
Another issue is data sparsity; the limited data availability impedes the model's generalization ability.
For example, in the publicly accessible Ali-CCP dataset, only about 3.85\% of displayed items received clicks, and merely 0.55\% of these clicks resulted in conversions (see Appendix~\ref{appendix:exp_details} for details).

Some recent studies have attempted to alleviate these problems by utilizing the entire data, including non-clicked samples.
ESMM \cite{Ma18Esmm} tackles these challenges by simultaneously predicting CTR and CVR across the entire space, developing distinct models for each task, and jointly optimizing them.
However, ESMM tends to overestimate CVR and CTCVR compared to the ground truth, which arises from its inability to accurately account for the causal relationship between click and conversion event \cite{Wang22Escm2}.
To address this issue, ESCM$^2$ \cite{Wang22Escm2} incorporates either an inverse propensity score (IPS) or a doubly robust (DR) regularizer into the training objective, demonstrating that this approach results in an unbiased CVR estimator.
However, ESCM$^2$ relies on scarce clicked samples when constructing the regularizer, making it challenging to ensure the unbiasedness of the CVR estimator across the entire space \cite{Zhu23Dcmt}.
To fully utilize the entire dataset, DCMT \cite{Zhu23Dcmt} introduces a counterfactual space for non-clicked samples that mirrors the corresponding factual space but with inverted conversion labels and subsequently incorporates this space into the model training.
Nonetheless, this approach naively assigns conversion labels of 1 to all non-clicked samples in the counterfactual space, leading to a distribution that significantly differs from that of the factual space.

Considering these factors, we devise a novel framework for CVR prediction that generates conversion labels for non-clicked samples through counterfactual inference.
Our main contributions are as follows:
\begin{itemize}[leftmargin=*]
    \item We develop a novel conversion label generation algorithm for non-clicked samples that leverages counterfactual inference, extending the knowledge of clicked samples to the non-clicked samples. 
    Specifically, we first obtain predicted CVR (pCVR) for non-clicked samples and transform the values into binary counterfactual conversion labels.
    The generated labels are utilized in the training process of a CVR prediction model.
    \item We perform experiments on both offline and online settings. We conduct extensive offline experiments on public datasets, as well as an online A/B test to demonstrate our proposed framework's superiority.
    The experimental results show that our approach improves upon the state-of-the-art methods by 1.01\% and 1.02\% on average in CVR and CTCVR AUC, respectively.
    Moreover, our approach achieves 17.35\% and 5.60\% improvement in online CVR and CTCVR over the state-of-the-art method, respectively, further validating the effectiveness of our algorithm.
    \item We analyze the performance of our approach on latent conversion data, which includes user data from the test set where users did not click during the training and validation phases but showed click behavior during the test phase. 
    The experimental results demonstrate that our approach achieves higher CVR and CTCVR estimation performance compared to baselines, ensuring robust performance in real-world scenarios where unseen data are common.
\end{itemize}

\section{Preliminaries}
\label{sec:preliminaries}
\subsection{Notation}
\label{subsec:notations}
A random variable and its realization are denoted by uppercase letters (e.g., $C$) and lowercase letters (e.g., $c$), respectively. 
Letters in calligraphic fonts, such as $\cC$, denote the sample space of the corresponding random variable, and $\bbP()$ represents the probability distribution of the random variable (e.g., $\bbP(C)$).
A lowercase letter with a hat notation (e.g., $\hat{c}$) denotes the predicted value of the corresponding realization.

\subsection{Problem Formulation}
\label{subsec:prob_form}
Let $\UserSpace = \{u_1, u_2, \dots, u_m\}$ denote the set of $m$ users and $\ItemSpace = \{i_1, i_2, $ $\dots, i_n\}$ denote the set of $n$ items. 
The exposure space $\ExposeSpace$ is a subset of  $\UserSpace \times \ItemSpace$, representing only the user-item pairs that are actually exposed. 
Additionally, we denote $\FeatRv_{u,i}$ as the feature of a user-item pair $(u,i)$.
The click, non-click, and conversion spaces are represented by $\ClickSpace$, $\NonClickSpace$, and $\ConvSpace$, respectively. 
Let $\ClickRv_{u,i}$ and $\ConvRv_{u,i}$ represent the occurrence of a click and conversion between user $u$ and item $i$, respectively.
Then, the spaces can be formulated as follows:
\begin{align}
    \ClickSpace
    &=
    \{(u,i) \in \ExposeSpace: c_{u,i}=1\},
    \label{def:click_space}
    \\
    \NonClickSpace
    &=
    \{(u,i) \in \ExposeSpace: c_{u,i}=0\},
    \label{def:nonclick_space}
    \\
    \ConvSpace
    &=
    \{(u,i) \in \ExposeSpace: v_{u,i}=1\}.
    \label{def:conv_space}
\end{align}

If $v_{u,i}$ is fully observed for $(u, i) \in \ExposeSpace$, the ideal loss function of CVR estimation can be formulated as follows:
\begin{align}
    \cL_{\text{ideal}}
    =
    \bbE_{(u,i) \in \ExposeSpace} [\ell(\hat{v}_{u,i}, v_{u,i})]
    = 
    \frac{1}{|\ExposeSpace|}\!
    \sum_{(u,i) \in \ExposeSpace}\!
    \ell(\hat{v}_{u,i}, v_{u,i}),
    \label{eq:ideal_cvr_loss}
\end{align}
where $\hat{v}_{u,i}$ denotes the pCVR.
The function $\ell(\cdot, \cdot)$ denotes the binary cross entropy loss, defined as $\ell(\hat{v}_{u,i}, v_{u,i}) = -v_{u,i}\log(\hat{v}_{u,i})-(1-v_{u,i})\log(1-\hat{v}_{u,i})$.
However, since the conversion labels are not fully observable across the entire space, a naive CVR estimator calculates the loss \textit{only} over $\ClickSpace$, formulated as follows,
\begin{align}
    \cL_{\text{naive}}
    =
    \bbE_{(u,i) \in \ClickSpace} [\ell(\hat{v}_{u,i}, v_{u,i})]
    = 
    \frac{1}{|\ClickSpace|}
    \sum_{(u,i) \in \ClickSpace}
    \ell(\hat{v}_{u,i}, v_{u,i}),
    \label{eq:naive_cvr_loss}
\end{align}
which is biased, i.e., $|\cL_{\text{ideal}} - \cL_{\text{naive}}| \neq 0$.
Consequently, previous studies \cite{Ma18Esmm, Wang22Escm2, Zhu23Dcmt} have focused on mitigating this bias in CVR estimation to improve performance.

\subsection{Prior Works and Their Limitations}
\label{subsec:prior_work}
We introduce prior works and discuss their limitations.
\subsubsection{ESMM \cite{Ma18Esmm}}
\label{subsubsec:prior_work_ESMM}
ESMM consists of the individual CTR and CVR models to predict both CTR and CVR over the entire space by jointly optimizing these tasks. 
The objective of ESMM is as follows:
\begin{align}
    \cL_{\text{ESMM}}
    &=
    \cL_{\text{CTR}} +
    \cL_{\text{CTCVR}},
    \label{eq:ESMM_loss}
\intertext{where}
    \cL_{\text{CTR}}
    &=
    \bbE_{(u,i) \in \ExposeSpace} [\ell(\hat{c}_{u,i}, c_{u,i})],
    \label{eq:ctr_loss}
    \\
    \cL_{\text{CTCVR}}
    &=
    \bbE_{(u,i) \in \ExposeSpace} [\ell(\hat{c}_{u,i}\hat{v}_{u,i}, c_{u,i}v_{u,i})].
    \label{eq:ctcvr_loss}
\end{align}
Here, $\hat{c}_{u,i}$ denotes the predicted CTR.
However, as mentioned in \cite{Wang22Escm2}, 
the CVR estimate from ESMM tends to be inherently higher than the ground truth because it assumes that the click and conversions are conditionally independent given user-item features.

\subsubsection{ESCM$^2$\cite{Wang22Escm2}}
\label{subsubsec:prior_work_ESCM$^2$}
ESCM$^2$ attempts to address the following question: ``Will the users convert on the recommended items if they click them?" by modeling CVR as $\bbP(V_{u,i}=1 \mid X, do (C_{u,i}=1))$, an interventional probability, where the ``$do$" denotes the \textit{do-operator} indicating an intervention.
To obtain this probability, ESCM$^2$ introduces a regularizer based on the IPS to remove the confounding bias stemming from covariates $X$, referred to as ESCM$^2$-IPS.
It demonstrates that training a model with this regularizer encourages the model to predict the desired CVR.
Furthermore, ESCM$^2$ leverages a new auxiliary imputation task trained over $\ExposeSpace$ to improve the debiasing performance in $\ClickSpace$ and the stability of the training process, also known as ESCM$^2$-DR.
The objective of each model can be formulated as follows:
\begin{align}
    \cL_{\text{ESCM}^2\text{-}Y}
    &=
    \cL_{\text{CTR}} +
    \cL_{\text{CTCVR}} +
    \alpha \cL_{\text{CVR}}^{Y},
    \label{eq:ESCM$^2$_loss} 
\intertext{
where $Y \in \{\text{IPS}, \text{DR}\}$ and $\alpha$ is a hyper-parameter. 
Each CVR loss is defined as follows:
}
    \cL_{\text{CVR}}^{\text{IPS}}
    &=
    \bbE_{(u,i) \in \ClickSpace} 
    \left[
    \frac{\ell(\hat{v}_{u,i}, v_{u,i})}{\hat{c}_{u,i}}
    \right],
    \label{eq:ESCM2_IPS_cvr_loss}
    \\
    \cL_{\text{CVR}}^{\text{DR}}
    &=
    \bbE_{(u,i) \in \ExposeSpace} 
    \Bigg[
    \hat{\delta}_{u,i}
    +
    \frac{c_{u,i} \big(\hat{e}_{u,i} + \hat{e}_{u,i}^2 \big)}{\hat{c}_{u,i}}
    \Bigg].
    \label{eq:ESCM2_DR_cvr_loss}
\end{align}
Here, $\hat{\delta}_{u,i} = \hat{\ell}(\hat{v}_{u,i}, v_{u,i})$ is a CVR estimation error and $\hat{e}_{u,i} = \ell(\hat{v}_{u,i}, v_{u,i}) - \hat{\ell}(\hat{v}_{u,i}, v_{u,i})$ is an error deviation.

Nonetheless, ESCM$^2$-IPS constructs the CVR loss solely based on observed clicked data without utilizing non-clicked data.
This approach does not fully resolve the selection bias in $\NonClickSpace$, as fundamental differences exist between clicked and non-clicked items (i.e., lack of overlap).
Furthermore, ensuring that the unbiased CVR estimation trained on $\ClickSpace$ (ESCM$^2$-DR) performs effectively on $\ExposeSpace$ remains challenging \cite{Zhu23Dcmt}.

\subsubsection{DCMT\cite{Zhu23Dcmt}}
\label{subsubsec:prior_work_dcmt}
DCMT defines the counterfactual space of $\NonClickSpace$, denoted by $\CounterNonClickSpace$, where each sample in this space is defined as the mirror image of the corresponding factual sample but with an inverted conversion label, e.g., $v_{u,i}^* = 1-v_{u,i}$. 
This implies that the conversion label is set to 1 for samples in $\CounterNonClickSpace$, as all the conversion labels in $\NonClickSpace$ are defined to be 0.
Then, the CVR loss is defined as:
\begin{align}
\hspace{-.25em}
    \cL_{\text{CVR}}^{\text{DCMT}}
    \!&=\!
    \bbE_{(u,i) \in \ClickSpace} 
    \!\left[
    \frac{\ell(\hat{v}_{u,i}, v_{u,i})}{\hat{c}_{u,i}}
    \right]
    \!+\!
    \bbE_{(u,i) \in \CounterNonClickSpace} 
    \left[
    \frac{\ell(\hat{v}^*_{u,i}, v^*_{u,i})}{1-\hat{c}_{u,i}}
    \right],
    \label{eq:dcmt_cvr_loss}
\end{align}
where $\hat{v}^*_{u,i}$ is the corresponding predicted counterfactual CVR.
Then, the final objective of DCMT is formulated as follows\footnote{The original paper integrates a novel regularizer into the final objective; however, we omit this term for the sake of conciseness.}:
\begin{align}
    \cL_{\text{DCMT}}
    =
    \cL_{\text{CTR}} +
    \cL_{\text{CTCVR}} +
    \alpha \cL_{\text{CVR}}^{\text{DCMT}}.
    \label{eq:dcmt_loss}
\end{align}
However, naively assuming all the counterfactual conversion labels to be 1 in $\CounterNonClickSpace$ is not optimal since the conversion label distribution of factual space $\ClickSpace$ will significantly differ from that of $\CounterNonClickSpace$.

\section{The \ours{} Framework}
\label{sec:ours}
\begin{figure}[t]
    \centering
    \hfill
    \begin{subfigure}{0.43\columnwidth}
        \centering
        \begin{tikzpicture}
        \node[circle,draw] (X) at (0,1) {$X$};
        \node[circle,draw] (C) at (-1,0) {$C$};
        \node[circle,draw] (V) at (1,0) {$V$};
        \node[circle,draw] (T) at (0,-1) {$T$};
        \node[overlay,circle,draw,dashed,inner sep=0.6mm] (Z) at (1.6,.75) {$Z$};
        \node[below=0mm of T] {\scriptsize Click \& Conversion};
        \node[below=0mm of C,xshift=-2mm] {\scriptsize Click};
        \node[overlay,below=0mm of V,xshift=2mm] {\scriptsize Conversion};
        \node[above=0mm of X] {\scriptsize User-item features};
        \node[overlay,above=0mm of Z] {\scriptsize Exogenous};
        \draw[->] (X) -- (C);
        \draw[->] (X) -- (V);
        \draw[->] (C) -- (V);
        \draw[->] (C) -- (T);
        \draw[->] (V) -- (T);
        \draw[->] (Z) -- (V);
        
        \end{tikzpicture}
        \caption{SCM of CVR prediction.}
        \label{fig:our_scm}
    \end{subfigure}
    \hfill
    \begin{subfigure}{0.43\columnwidth}
        \centering
        \begin{tikzpicture}
        \node[circle,draw] (X) at (0,1) {$X$};
        \node[circle,draw] (C) at (-1,0) {$1$};
        \node[circle,draw] (V) at (1,0) {$V$};
        \node[circle,draw] (T) at (0,-1) {$T$};
        \node[overlay,circle,draw,dashed,inner sep=0.6mm] (Z) at (1.6,.75) {$Z$};
        \node[opacity=0,below=0mm of T] {\scriptsize Click \& Conversion};
        \node[opacity=0,above=0mm of X] {\scriptsize User-item features};
        \draw[->] (X) -- (V);
        \draw[->] (C) -- (V);
        \draw[->] (C) -- (T);
        \draw[->] (V) -- (T);
        \draw[->] (Z) -- (V);
        \end{tikzpicture}
        \caption{Intervention on $C$.}
        \label{fig:action}
    \end{subfigure}\hfill\null

    \medskip

    \begin{subfigure}{\columnwidth}
        \centering
        \includegraphics[width=\columnwidth]{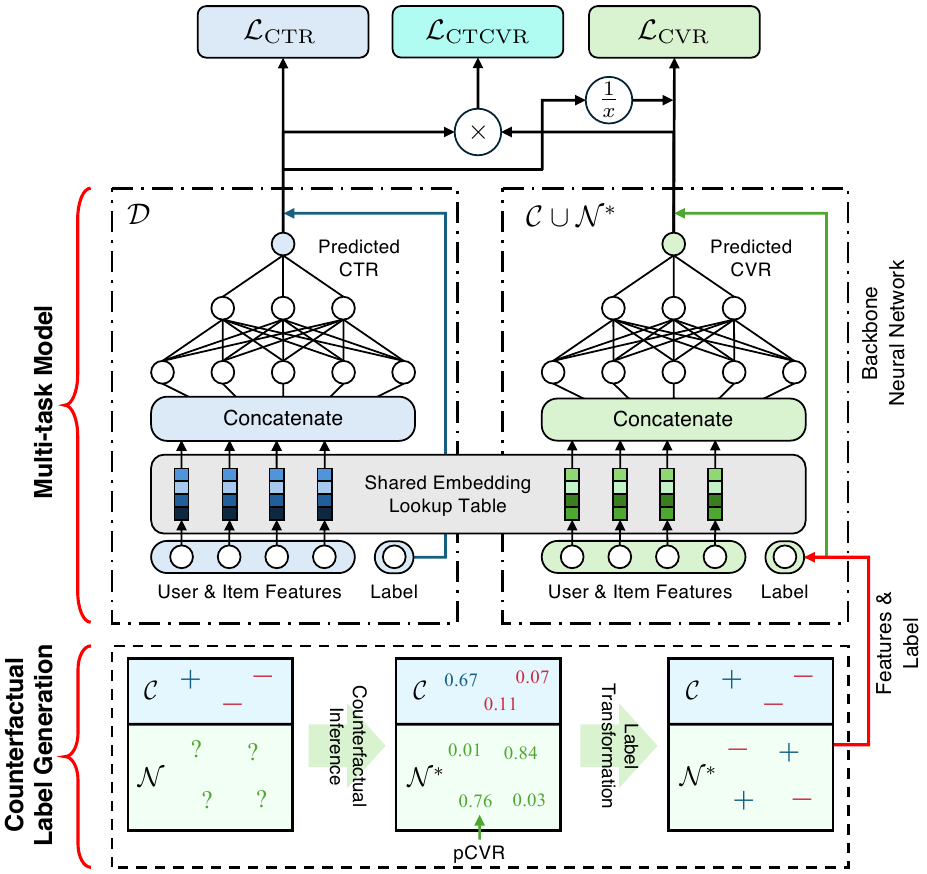}
        \caption{Illustration of \ours{}.}
        \label{fig:main_architecture}
    \end{subfigure}

    \caption{Overall procedure of \ours{}. (a) $Z$ represents the exogenous variable for the conversion event, acting as an underlying factor of user conversion behavior. (b) The hypothetical intervention on $C$ simulates a scenario where the click event is set to 1 to generate a counterfactual conversion outcome. (c) Counterfactual labels for non-clicked samples are generated through counterfactual inference and label transformation (bottom). The generated labels are then utilized to train a CVR prediction model (top).}
    \label{fig:our_method}
\end{figure}
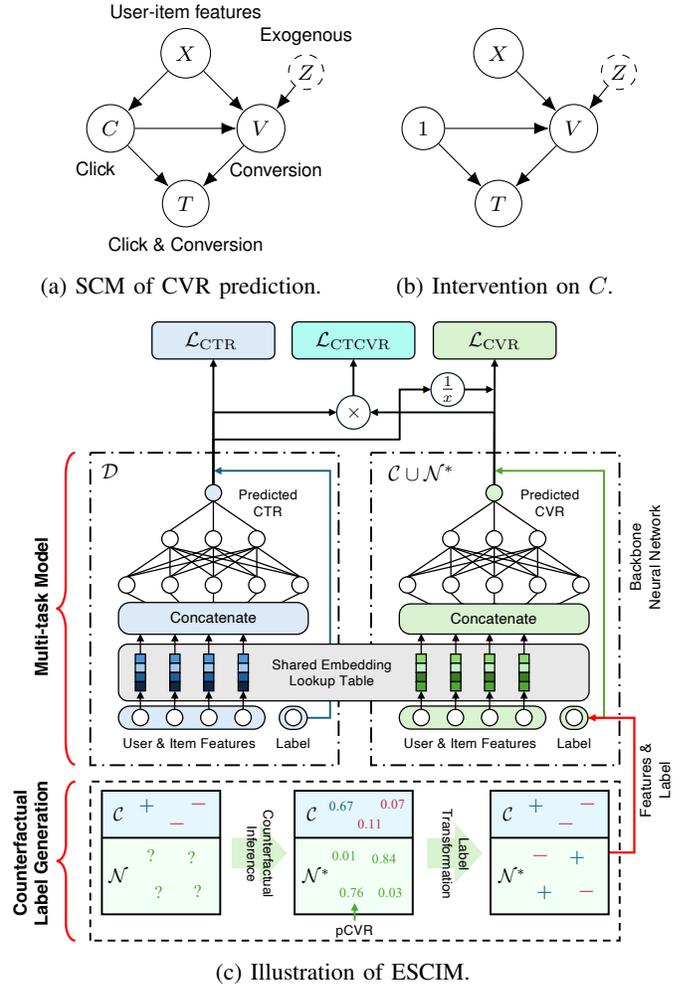

We describe our framework for CVR prediction, which incorporates counterfactual inference to utilize the \textit{non-clicked} dataset effectively.
Within a trained causal model reflecting real-world user behavior, we apply a hypothetical intervention on the click event $C$ ($do(C=1)$) for non-clicked samples to generate counterfactual conversion labels, denoted by $v^*$.
This action implies that, within an imaginary scenario, all users are compelled to click on the exposed item and decide whether to convert on the item.
This procedure creates a counterfactual space for the non-clicked space $\NonClickSpace$, denoted by $\CounterNonClickSpace$.
Subsequently, we incorporate the generated space $\CounterNonClickSpace$ into the training of a CVR prediction model.
The procedure of our method is illustrated in \cref{fig:our_method}.

\subsection{Counterfactual Inference}
Our primary objective is to estimate a counterfactual probability:
\[
    \bbP\left(V_{u,i}(C_{u,i}=1)=1 \mid X_{u,i}, C_{u,i}=0, V_{u,i}=0\right),
\]
which represents the probability that user $u$ would have converted on item $i$ if the user had clicked it, given that the user $u$ did not click on item $i$ ($C_{u,i}=0$) and therefore did not convert ($V_{u,i}=0$). 
Here, $V_{u,i}(C_{u,i}=1)$ denotes the potential outcome of a conversion event that would have been observed if $C_{u,i}$ had been set to 1.
\label{subsec:ci}

We adopt Pearl's causal framework \cite{Pearl16Ci} to perform counterfactual inference. The aforementioned counterfactual quantity is theoretically implausible to identify \cite{Correa21Nested}, meaning that the provided distribution $P(X,C,V)$ is causally insufficient to determine the quantity, regardless of the dataset size.\footnote{Assumptions encoded in the graph prohibit us from identifying the counterfactual quantity. However, under determinism, the condition $V_{u,i}=0$ can indeed be dropped, finally yielding the term $P(V_{u,i}=1 \mid X_{u,i}, C_{u,i}=1)$. Yet ignorability may not hold, and inferring $Z$ from the clicked and non-converted cases would allow us to partly infer the characteristics of the non-clicked. To acknowledge the differences between clicked and non-clicked, we should exercise prior—yet uncertain—knowledge conservatively, as we did when transforming pCVRs into hard labels.}
Consequently, we resort to estimating the quantity by simulating one of many plausible values. This process involves constructing a structural causal model (SCM) compatible with the given $P(X,C,V)$ and performing hypothetical interventions to sample counterfactual outcomes.

We represent the CVR prediction problem through an SCM, as depicted in \cref{fig:our_scm}\footnote{In drawing a causal graph, it is convention to omit exogenous variables. Here, we include the exogenous variable for $V$, which is critical in understanding the counterfactual inference employed in this paper.}, which explains the following points: 1) $\FeatMat \rightarrow \ClickMat$ indicates that the click event depends on user-item features (an intervention on $\ClickMat$ will sever the connection as shown in \cref{fig:action});
2) $\FeatMat, \ClickMat, Z \rightarrow \ConvMat$ demonstrates that the conversion event is defined after the click event, where $Z$ denotes the exogenous variable for the conversion event, which serves as a hidden factor that determines the user conversion behavior.
3) $\ClickMat, \ConvMat \rightarrow T$ illustrates that a click \& conversion event, represented as $T$.

The counterfactual probability can be estimated through the following procedure: ``Abduction", ``Action", and ``Prediction". In the ``Abduction" step, we infer the posterior of the exogenous variable of the conversion event, based on the observation.
In the ``Action" step (\cref{fig:action}), we remove all the incoming edges to a click event $C$ by applying a hypothetical intervention and setting its value to 1 ($do(C=1)$), which virtually imposes all users to click all exposed items.
Finally, in the ``Prediction" step, we predict the potential outcomes of the counterfactual scenarios based on the modified values reflecting a counterfactual world.
Since we are only interested in $V$, we model the causal mechanism for $V$ with $X$, $C$, and $Z$ by building a multi-layer perceptron (MLP) model $f_{\theta}(\cdot)$ and training the model with observed data, which will be used in the ``Prediction" step.
We denote this step as ``Pre-training".

\subsubsection{Step 1: Pre-training}
\label{subsubsec:pre_training}
To facilitate counterfactual inference, the objective of this stage is constructing and training a conversion model $f_{\theta}(\cdot)$  utilizing observational data together with an exogenous variable. We assume that an exogenous variable onto conversion $V$, i.e., $Z$, is drawn from the standard Gaussian distribution $\mathcal{N}(\mathbf{0}, \mathbf{I})$, following \cite{Yang21topn, Wang22Cor}. 

With sampled exogenous values, we compile a dataset by gathering tuples $(x_{u,i}, c_{u,i}, z_{u,i}, v_{u,i})$ for $(u,i) \in \ClickSpace$.
Then, we train the MLP model $f_{\theta}(\cdot)$ with the constructed dataset, where the training loss is defined as
\begin{align}
    \cL_{\text{pre-train}}
    =
    \bbE_{(u,i) \in \ClickSpace} 
    \left[
    \ell(f_{\theta}(x_{u,i}, z_{u,i}), v_{u,i})
    \right].
    \label{eq:pre_train_loss}
\end{align}

\begin{algorithm}[t] 
    \SetKwInOut{KwIn}{Input}
    \SetKwInOut{KwOut}{Output}

    \KwIn{Observed clicked data $(x_{u,i}, 1, v_{u,i})$ for $(u,i) \in \ClickSpace$, $(x_{u,i}, 0)$ for $(u,i) \in \NonClickSpace$, MLP $f_{\theta}(\cdot)$, and VAE $g_{\phi}(\cdot).$}
    \KwOut{The predicted counterfactual CVR
    $\hat{v}^*_{u,i}$ for $(u,i) \in \CounterNonClickSpace$}\medskip
    
    \textbf{Pre-training}: Draw $z_{u,i}$ from $\mathcal{N}(\mathbf{0} , \mathbf{I})$ for $(u,i) \in \ClickSpace$. Train $f_{\theta}(\cdot)$ with the observed clicked data and $z_v$, where the loss function is  $\cL_{\text{pre-train}}$.\smallskip

    \textbf{Abduction}: Train $g_{\phi}(\cdot)$ using the observed click data and $z$, with the ELBO objective (Eq.~\ref{eq:elbo_vae}).\smallskip
    
    \textbf{Action}: Apply an intervention on click events, e.g.,  $do(C_{u,i}=1)$, for $(u,i) \in \NonClickSpace$.\smallskip
    
    \textbf{Prediction}: Draw $z_{u,i}$ from the output of the trained encoder of $g_{\phi}(\cdot)$, and compute 
    $\hat{v}^*_{u,i} = f_{\theta}(x_{u,i}, z_{u,i})$ for $(u,i) \in \NonClickSpace$.

    \caption{Counterfactual Inference}
    \label{algo:counterfactual_inference}
\end{algorithm}

\subsubsection{Step 2: Abduction}
\label{subsubsec:abduction}
This step aims to acquire the posterior distribution $p(z \mid X, C, V)$, which is pivotal for subsequent steps.
However, due to its complexity, directly sampling from the posterior distribution is challenging.
Thus, we leverage the variational inference \cite{Blei17}, which approximates $p(Z \mid X, C, V)$ to a Gaussian distribution $q(z) = \mathcal{N}(z \mid \mu, \sigma^2)$, where $\mu$ and $\sigma$ are learnable parameters. The latent random variable is then drawn according to $Z \sim q$.
To perform this, we construct a variational autoencoder (VAE) $g_{\phi}(\cdot)$ and train it with the clicked data.
The VAE provides a structured framework to learn the underlying latent representation, which captures both observed and unobserved influences, including exogenous variables.
To facilitate the back-propagation of gradients through sampling operations, we utilize the reparameterization trick \cite{Kingma14, Liang20}.
Furthermore, Kullback–Leibler divergence (KLD) annealing \cite{Liang20} is also employed to mitigate the risk of posterior collapse.
Thus, the evidence lower bound (ELBO) can be written as follows:
\begin{align}
    \bbE_{(u,i) \in \ClickSpace}
    \!\left[
    \bbE_{q_{\phi}(z)}\!
    \left[
    p(z \mid x_{u,i}, v_{u,i})
    \right]
    \!-\! 
    \beta
    \text{KL}(p(z)|| q_{\phi}(z))
    \right],
    \label{eq:elbo_vae}
\end{align}
where $\beta$ is a hyperparameter that regularizes the KLD.

\subsubsection{Step 3: Action}
\label{subsubsec:action}
As in \cref{fig:action}, we apply a hypothetical intervention on the click event by setting the value to 1, $\textit{do}(C_{u,i}=1)$ for $(u,i) \in \NonClickSpace$.
This action generates the counterfactual space $\CounterNonClickSpace$, where all click events are forced to be 1.

\subsubsection{Step 4: Prediction}
\label{subsubsec:prediction}
In this step, we compute the counterfactual CVR for samples in $\CounterNonClickSpace$ by feeding $\{(x_{u,i}, z_{u,i})\}$ into the pre-trained model $f_{\theta}(\cdot)$, where $z$ is sampled from the approximated distribution $q_{\phi}(z)$, obtained in the ``Abduction" step.
The output is a pCVR for each counterfactual example, which is denoted by $\hat{v}^{*}_{u,i}$.
We summarize the complete learning process in \cref{algo:counterfactual_inference}.

\subsection{How to transform the predicted counterfactual CVR to a hard label?}
\label{subsec:convert_labels}
As a consequence of the preceding steps, we derive the pCVR for each counterfactual example in $\CounterNonClickSpace$.
The subsequent task is to transform these counterfactual pCVRs into binary conversion labels (0, 1) for training purposes.
We pursue this transformation through two methods: 1) a \textit{max approach} and 2) a \textit{ratio approach}.
Both methods rely on the true distribution of CVR in $\ClickSpace$, assuming similarity in user behavior between $\CounterNonClickSpace$ and $\ClickSpace$, thus enabling the comparability of CVR distributions in both spaces.
This assumption aligns with the goal of counterfactual inference, which aims to enhance our understanding of causal relationships by predicting outcomes in unobserved scenarios and simulating alternative possibilities based on existing knowledge.

\begin{algorithm}[t] 
    \SetKwInOut{KwIn}{Input}
    \SetKwInOut{KwOut}{Output}

    \KwIn{Predicted counterfactual CVR
    $\hat{v}^*_{u,i}$ for $(u,i) \in \CounterNonClickSpace$, $(x_{u,i}, c_{u,i})$ for $(u,i) \in \ClickSpace$, Trained MLP $f_{\theta}(\cdot)$, Trained VAE $g_{\phi}(\cdot)$}
    \KwOut{The binary counterfactual conversion label
    $v^*_{u,i}$ for $(u,i) \in \CounterNonClickSpace$}
    \For{$(u,i) \in \ClickSpace$}{
        Sample $z_{u,i}$ from the trained Encoder $g_{\phi}(\cdot)$ \\
        Obtain $\hat{v}_{u,i} = f_{\theta}(x_{u,i}, z_{u,i})$
    }
    
    \For{$(u,i) \in \NonClickSpace$}{
        Label the sample as $v^*_{u,i}=1 \text{ \textbf{if} }\hat{v}^*_{u,i} \geq \max_{(u,i) \in \ClickSpace} \hat{v}_{u,i} \text{ \textbf{else} }0$
    }

    \caption{A Max Approach}
    \label{algo:convert_labels_max}
\end{algorithm}
\begin{algorithm}[t] 
    \SetKwInOut{KwIn}{Input}
    \SetKwInOut{KwOut}{Output}

    \KwIn{The predicted counterfactual CVR
    $\hat{v}^*_{u,i}$ for $(u,i) \in \NonClickSpace$}
    \KwOut{The binary counterfactual conversion label
    $v^*_{u,i}$ for $(u,i) \in \CounterNonClickSpace$}
        
    $k \leftarrow |\ConvSpace| \times |\NonClickSpace| \mathbin{/} |\ClickSpace|$; $\cS \leftarrow \varnothing $; $\textit{idx} \leftarrow 0$
    
    \For{$(u,i) \in \NonClickSpace$}{
        \eIf{\text{idx} $< k$}{
            $\cS \leftarrow \cS \cup (u,i)$        
        }
        {$(\tilde{u}, \tilde{i}) \gets \arg\min_{(u,i)} \hat{v}_{u,i}$ 
        
        \If{$\hat{v}_{u,i} > \hat{v}_{\tilde{u}, \tilde{i}}$}{
            $\cS \leftarrow \cS \setminus \{(\tilde{u}, \tilde{i})\} \cup \{(u,i)\}$
        }
        }
        $\textit{idx} \leftarrow \textit{idx} + 1$
    }
    
    \For{$(u,i) \in \NonClickSpace$}{
    Label the sample as $v^*_{u,i}=1$ \textbf{ if } $(u,i) \in \cS$ \textbf{ else } 0
    }
    \caption{A Ratio Approach}
    \label{algo:convert_labels_ratio}
\end{algorithm}
\subsubsection{A max approach}
\label{subsec:approach_thr}
We identify non-clicked samples with a high likelihood of conversion using the maximum \textit{pCVR} from clicked samples as a threshold, labeling non-clicked samples with pCVRs exceeding this threshold as conversions.
Firstly, for each sample in $\ClickSpace$, we obtain the pCVR by feeding its data to the trained network $f_{\theta}(\cdot)$.
Then, if $\hat{v}^*_{u,i} \geq \max_{(u,i) \in \ClickSpace} \hat{v}_{u,i}$ for $(u,i) \in \CounterNonClickSpace$, we label the sample as 1 ($v^*_{u,i}=1$); otherwise, we label it as 0 ($v^*_{u,i}=0$).
The rationale for using the $\max$ function stems from the uncertainty inherent in the counterfactual space.
Given that we train $f_{\theta}(\cdot)$ with very scarce clicked data and infer the pCVRs for a substantial amount of non-clicked data, there is a possibility of incorrect predictions.
Therefore, we only assign a conversion label of 1 to samples with a pCVR that exceeds the maximum of all factual pCVRs.
We summarize the complete procedure in \cref{algo:convert_labels_max}.
Our model that utilizes this approach is denoted by \ours-max.

\subsubsection{A ratio approach}
\label{subsec:approach_topk}
We assign conversion labels of 1 to the non-clicked samples with top-$k$ pCVRs to yield the same conversion ratio as observed in the clicked samples.
Specifically, we first estimate the number of conversions in $\CounterNonClickSpace$, denoted by $k$, by calculating the number of conversions in $\ClickSpace$ and scaling it by the ratio of the size of $\CounterNonClickSpace$ to $\ClickSpace$.
Next, we identify the indices corresponding to the top-$k$ values of the counterfactual pCVRs.
Finally, we assign a label of 1 to these top-$k$ samples and 0 to all others. 
We provide a detailed summary of the procedure in \cref{algo:convert_labels_ratio}.
Our model that employs this approach is referred to as \ours-ratio.

\subsection{Objective of \ours{}}
\label{subsec:overall_loss}
We first construct a multi-task model, such as ESMM, which consists of CTR and CVR prediction models, as shown in \cref{fig:main_architecture}.
Then, we train the model using the factual click space $\ClickSpace$, the non-click space $\NonClickSpace$, and the generated counterfactual space $\CounterNonClickSpace$ with the following objective: 
\begin{align} 
    \cL_{\text{\ours}}
    =
    \cL_{\text{CTR}} + \cL_{\text{CTCVR}} + \alpha_{F} \cL_{\text{CVR-F}} + \alpha_{CF} \cL_{\text{CVR-CF}},\phantom{....}
     \raisetag{12pt} \label{eq:obj_ours}
\end{align}
where $\cL_{\text{CTR}}$ and $\cL_{\text{CTCVR}}$ are defined in \cref{eq:ctr_loss,eq:ctcvr_loss}, respectively.
$\cL_{\text{CVR-F}}$ and $\cL_{\text{CVR-CF}}$ are factual and counterfactual CVR loss, respectively, which are defined as follows:
\begin{align}
    \cL_{\text{CVR-F}}
    &=
    \bbE_{(u,i) \in \ClickSpace} 
    \left[
    \frac{\ell(\hat{v}_{u,i}, v_{u,i})}{\hat{c}_{u,i}}
    \right], \\
    \cL_{\text{CVR-CF}}
    &=
    \bbE_{(u,i) \in \CounterNonClickSpace} 
    \left[
    \frac{\ell(\hat{v}_{u,i}, v_{u,i}^*)}{\hat{c}_{u,i}}
    \right].
    \label{eq:obj_ours_losses}
\end{align}
$\alpha_{F}$ and $\alpha_{CF}$ are weight parameters for factual and counterfactual CVR loss, respectively.
If $\alpha_{CF}=0$, the objective becomes equivalent to that of ESCM$^2$.

\section{Experiments}
\label{sec:exp}

\begin{table*}
\caption{Performance comparisons of the proposed model (left) with baselines on the Ali-CCP dataset across various backbones, and (right) with baselines on the Ali-Express dataset.
For Ali-Express, the backbone model is an MLP with [512, 256, 128] layers. The best results are in bold, and the previous best-performing baseline results are underlined.}
\label{table:perf_comp_on_datasets}
\adjustbox{max width=\textwidth}{
\begin{tabular}{@{}lcccccccc@{}}
\toprule 
& \multicolumn{4}{c}{\textbf{Ali-CCP across backbones}} & \multicolumn{4}{c}{\textbf{Ali-Express w/ MLP across countries}} \\
\cmidrule(r){2-5} \cmidrule(l){6-9} 
\textbf{Method} & MLP & DeepFM & AutoInt & DCN-V2 &  AE-ES & AE-FR & AE-NL & AE-US \\
\midrule 
ESMM          & 0.6474 / 0.6379 & 0.6398 / 0.6213 & 0.6543 / 0.6302 & 0.6495 / 0.6373 &    0.8099 / 0.8717 & 0.7949 / 0.8500 & 0.7795 / 0.8479 & 0.7952 / 0.8425 \\
Multi-IPS     & 0.6523 / 0.6390 & 0.6431 / 0.6244 & 0.6587 / 0.6357 & 0.6535 / 0.6379 &   0.8205 / 0.8592 & 0.7947 / 0.8504 & 0.7829 / 0.8377 & 0.8068 / 0.8433 \\
Multi-DR      & 0.6437 / 0.6305 & 0.6388 / 0.6199 & 0.6542 / 0.6304 & 0.6488 / 0.6353 &  0.8135 / 0.8600 & 0.7943 / 0.8538 & 0.7803 / 0.8292 & 0.7956 / 0.8396 \\
ESCM$^2$-IPS  & 0.6691 / 0.6424 & 0.6487 / 0.6281 & 0.6681 / \underline{0.6407} & \underline{0.6609} / \underline{0.6374} &    0.8199 / 0.8777 & 0.8078 / 0.8511 & 0.7867 / 0.8511 & 0.8018 / 0.8573 \\
ESCM$^2$-DR   & 0.6609 / 0.6353 & 0.6502 / 0.6280 & 0.6485 / 0.6329 & 0.6489 / 0.6313 &   0.8112 / 0.8739 & 0.8073 / 0.8535 & 0.7772 / 0.8473 & 0.7917 / 0.8404 \\
DCMT          & \underline{0.6743} / \underline{0.6451} & \underline{0.6528} / \underline{0.6338} & \underline{0.6702} / 0.6389 & 0.6603 / 0.6350 &   \underline{0.8251} / \underline{0.8838} & \underline{0.8089} / \underline{0.8628} & \underline{0.7897} / \underline{0.8525} & \underline{0.8140} / \underline{0.8620} \\
ESCIM-max     & \textbf{0.6792} / 0.6487 & \textbf{0.6585} / \textbf{0.6401} & 0.6737 / 0.6413 & \textbf{0.6698} / 0.6467 &    \textbf{0.8314} / 0.8935 & \textbf{0.8210} / 0.8776 & \textbf{0.7951} / \textbf{0.8613} & \textbf{0.8332 / 0.8782} \\
ESCIM-ratio    & 0.6756 / \textbf{0.6566} & 0.6523 / 0.6382 & \textbf{0.6774} / \textbf{0.6475} & 0.6683 / \textbf{0.6534} &    0.8305 / \textbf{0.8951} & 0.8192 / \textbf{0.8821} & 0.7907 / 0.8583 & 0.8201 / 0.8643 \\
\bottomrule
\end{tabular}
}
\end{table*}
\begin{table*}[t]
\caption{Standard deviations of the test performance reported in Table I.}
\label{table:perf_comp_on_datasets_std}
\adjustbox{max width=\textwidth}{
\begin{tabular}{@{}lcccccccc@{}}
\toprule 
& \multicolumn{4}{c}{\textbf{Ali-CCP across backbones}} & \multicolumn{4}{c}{\textbf{Ali-Express w/ MLP across countries}} \\
\cmidrule(r){2-5} \cmidrule(l){6-9} 
\textbf{Method} & MLP & DeepFM & AutoInt & DCN-V2 &  AE-ES & AE-FR & AE-NL & AE-US \\
\midrule 
ESMM          & 0.0012 / 0.0014 & 0.0018 / 0.0012 & 0.0010 / 0.0010 & 0.0011 / 0.0013 &    0.0007 / 0.0008 & 0.0008 / 0.0006 & 0.0007 / 0.0008 & 0.0006 / 0.0007 \\
Multi-IPS     & 0.0029 / 0.0027 & 0.0031 / 0.0026 & 0.0030 / 0.0029 & 0.0029 / 0.0028 &   0.0020 / 0.0018 & 0.0018 / 0.0019 & 0.0014 / 0.0013 & 0.0015 / 0.0013 \\
Multi-DR      & 0.0026 / 0.0023 & 0.0029 / 0.0027 & 0.0019 / 0.0016 & 0.0017 / 0.0017 &   0.0014 / 0.0011 & 0.0013 / 0.0013 & 0.0010 / 0.0006 & 0.0009 / 0.0010 \\
ESCM$^2$-IPS  & 0.0021 / 0.0027 & 0.0026 / 0.0023 & 0.0029 / 0.0024 & 0.0021 / 0.0013 &    0.0012 / 0.0013 & 0.0013 / 0.0005 & 0.0012 / 0.0009 & 0.0011 / 0.0009 \\
ESCM$^2$-DR   & 0.0020 / 0.0024 & 0.0012 / 0.0020 & 0.0015 / 0.0008 & 0.0023 / 0.0008 &   0.0013 / 0.0011 & 0.0014 / 0.0010 & 0.0008 / 0.0007 & 0.0010 / 0.0011 \\
DCMT          & 0.0012 / 0.0021 & 0.0020 / 0.0018 & 0.0010 / 0.0007 & 0.0015 / 0.0011 &   0.0014 / 0.0011 & 0.0013 / 0.0011 & 0.0009 / 0.0006 & 0.0009 / 0.0010 \\
ESCIM-max     & 0.0017 / 0.0016 & 0.0017 / 0.0011 & 0.0010 / 0.0012 & 0.0010 / 0.0012 &    0.0015 / 0.0013 & 0.0007 / 0.0007 & 0.0008 / 0.0007 & 0.0010 / 0.0009 \\
ESCIM-ratio    & 0.0015 / 0.0017 & 0.0012 / 0.0014 & 0.0011 / 0.0012 & 0.0014 / 0.0015 &   0.0012 / 0.0014 & 0.0011 / 0.0013 & 0.0014 / 0.0012 & 0.0015 / 0.0014 \\
\bottomrule
\end{tabular}
}
\end{table*}

We conduct experiments to evaluate the performance of \ours{} and investigate the following research questions:
\begin{itemize}[leftmargin=*]
    \item \textbf{RQ1}: How does \ours{} perform on the CVR and CTCVR prediction tasks compared to the baselines across various backbones?
    \item \textbf{RQ2}: Does our model outperform the baseline in an online environment?
    \item \textbf{RQ3}: How does the performance of \ours{} vary across different thresholds in the label transformation step?
    \item \textbf{RQ4}: How significantly does accurately estimating the posterior distribution $p(Z \mid X,C,V)$ improve performance?
    \item \textbf{RQ5}: Does our model predict CVR and CTCVR more accurately on latent conversion data?
\end{itemize}

\subsection{Experimental Settings}
\label{subsec:exp_setting}

\subsubsection{Datasets}
\label{subsubsec:datasets}
We conduct experiments on two benchmark datasets, Ali-CCP and Ali-Express, both widely used in CVR and CTCVR prediction tasks \cite{Wang22Escm2, Zhu23Dcmt}.
The statistics of the datasets are detailed in Appendix~\ref{appendix:exp_details}. 
For both datasets, we randomly sample 10\% of the training set to create our validation dataset.
\begin{itemize}[leftmargin=*]
    \item \textbf{Ali-CCP}\footnote{\url{https://tianchi.aliyun.com/dataset/dataDetail?dataId=408}}: The dataset contains real-world traffic logs from the recommender system of Taobao, a Chinese online shopping platform.
    This dataset includes detailed logs of user interactions such as clicks and conversions, along with associated features like user demographics, item attributes, and contextual information.
    \item \textbf{Ali-Express}\footnote{\url{https://tianchi.aliyun.com/dataset/dataDetail?dataId=74690}} \cite{Li20}: The dataset comprises real-world search logs and user interaction data collected from AliExpress, a global online retail service based in China.
    Data are collected from four countries: Spain, France, Netherlands, and the United States.\footnote{Due to the substantial file size, Russia has been excluded from this experiment.}
    Each country's dataset is denoted by ``AE-(country code)", e.g., AE-ES for Spain.
\end{itemize}

\subsubsection{Baselines}
\label{subsubsec:baseline}
We compare our model with the following baselines, where all the models leverage a multi-task learning framework to enhance the CVR prediction performance: 1) ESMM~\cite{Ma18Esmm}; 2) Multi-IPS, Multi-DR~\cite{Zhang20Multi}; 3) ESCM$^2$-IPS, ESCM$^2$-DR~\cite{Wang22Escm2}; 4) DCMT~\cite{Zhu23Dcmt}.
Details of these models are explained in \cref{subsec:prior_work}.

\subsubsection{Backbones}
\label{subsubsec:backbone}
We experiment on different backbones for CTR and CVR models: 1) \textbf{MLP}: a traditional fully-connected neural network; 2) \textbf{DeepFM} \cite{Guo17Deepfm}: a model incorporating the factorization machines and deep neural networks; 3) \textbf{AutoInt} \cite{Song19Autoint}: a model employing multi-head self-attention to learn high-order feature interactions automatically; 4) \textbf{DCN-V2} \cite{Wang21Dcnv2}: a model utilizing deep and cross networks to learn effective explicit and implicit feature interactions efficiently.

\subsubsection{Parameter Settings}
\label{subsubsec:param_setting}
The embedding size for each feature is set to 5 for the Ali-CCP dataset and 32 for the Ali-Express dataset.
The dimension of $z$ is set to the total size of the embedding, which is (embedding size) $\times$ (number of features).
We set the batch size to 8,192.
The objective weight $\alpha_{F}$ is set to 0.1, as in \cite{Wang22Escm2}, and $\alpha_{CF}$ is set to 1e-4.
We implement $f_{\theta}(\cdot)$ using an MLP with $[512, 256, 128]$ layers.
For $g_{\phi}(\cdot)$, we use an MLP with $[512, 256, 128]$ layers for the encoder and $[128, 256, 512]$ layers for the decoder.
Further details regarding the backbones are provided in Appendix~\ref{appendix:imp_details}.

\subsubsection{Evaluation}
\label{subsubsec:evaluation}
Following prior work~\cite{Ma18Esmm, Wang22Escm2, Zhu23Dcmt}, we evaluate  both CVR and CTCVR prediction with the area under the ROC curve (AUC), and report the average performance over five runs with different random seeds per backbone.

\subsection{Performance Comparisons (RQ1)}
\label{subsec:rq1}
We report the mean and standard deviation of test performance for both the baselines and our models on the Ali-CCP and Ali-Express datasets in \cref{table:perf_comp_on_datasets,table:perf_comp_on_datasets_std}, respectively.
The metrics on the left and right of the slash represent the CVR AUC and the CTCVR AUC, respectively.
In general, our models consistently outperform the baselines in all backbones.
We have the following observations:
\begin{itemize}[leftmargin=*]
    \item In the Ali-CCP dataset, we observe that the \ours-max achieves the highest CVR AUC across all backbones except for AutoInt, showing a significant improvement of 0.87\% on average.
    \ours-max records the highest CVR AUC performance at 0.6792 using MLP, surpassing the existing state-of-the-art performance by 0.0049.
    Furthermore, except for DeepFM, \ours-ratio achieves the highest CTCVR AUC across all other backbones, demonstrating a significant improvement of 1.51\% on average.
    \item In the Ali-Express dataset, we observe that \ours-max achieves the highest CVR AUC across all countries.
    It also achieves the highest CTCVR AUC in the AE-NL and AE-US datasets, showing significant improvements of 1.03\% and 1.88\% over the best-performing baseline, respectively. 
    Similarly, \ours-ratio achieves the highest CTCVR AUC in the AE-ES and AE-FR datasets, with improvements of 1.28\% and 2.24\%, respectively.
    \item We believe that the superior performances on both datasets are attributed to the more accurate generation of counterfactual conversion labels for each data sample in the non-clicked space, closely reflecting the unknown true click behavior. 
    As a result, \ours{} demonstrates improved performance over the previous best model, DCMT, which simply sets all labels to 1.
\end{itemize}

\subsection{Online A/B Test (RQ2)}
\label{subsec:rq2}

\begin{table}[t]
    \centering
    \caption{Results of online A/B Test. Metrics with an upward pointing arrow indicate that the higher values are better, while metrics with a downward-pointing arrow indicate that lower values are preferable.}
    \label{table:online_results}
    \begin{adjustbox}{width=\columnwidth}
        \begin{tabular}{@{}rrrrrr@{}}
            \toprule
            \textbf{Metric} & \textbf{Day 1} & \textbf{Day 2} & \textbf{Day 3} & \textbf{Day 4} & \textbf{Day 5} \\
            \midrule
            CVR ($\uparrow$) & +21.55\% & +16.94\% & +12.59\% & +17.42\% & +18.24\% \\
            CTCVR ($\uparrow$) & +7.41\% & +4.76\% & +3.98\% & +5.56\% & +6.31\% \\
            CPA ($\downarrow$) & -27.16\% & -26.67\% & -16.05\% & -22.53\% & -24.38\% \\
            \bottomrule
        \end{tabular}
    \end{adjustbox}
\end{table}

To examine the performance of \ours{} in real-world applications, we conduct an online A/B test on our demand-side platform.
We train the model with the industrial dataset, which consists of records from our company's recommendation system (see Appendix~\ref{appendix:dataset_description} for details).
We choose \ours{}-max as our model and ESCM$^2$-IPS as a baseline, which shows competitive results to \ours{} in offline experiments.
We adopt an MLP as a backbone, with its parameters detailed in Appendix~\ref{appendix:exp_details}.
We randomly split users into two groups (buckets) using our online A/B test platform, ensuring that the characteristics and the number of users in the groups are comparable.
The online test was conducted over 5 days.

In addition to the aforementioned metrics, we consider cost-per-action (CPA), a crucial measure of marketing effectiveness. 
CPA is calculated as the total advertising spending divided by the number of conversions, which serves as a key indicator of the aggregate cost to acquire one paying customer at the campaign level. 
As the accuracy of CVR prediction improves, CPA typically decreases because higher prediction accuracy allows advertisers to target their campaigns more effectively toward users who are more likely to convert.

The results of the online A/B test are presented in \cref{table:online_results}.
Metrics with an upward-pointing arrow indicate that higher values are better, while metrics with a downward-pointing arrow indicate that lower values are preferable.
\ours{} improves the baseline by 17.35\% for CVR, 5.60\% for CTCVR, and 23.36\% for CPA on a macro-average over 5 days, showcasing the effectiveness of \ours{} in online environments.

\subsection{Analysis of Label Transformation (RQ3)}
\label{subsec:rq3}
To validate the effectiveness of the proposed label transformation strategies, we measure the test performance of \ours{} by varying the threshold for assigning a label of 1. 
Specifically, we begin by labeling samples with the highest pCVRs as 1 and continue this process in a descending order. 
Instead of reporting the actual threshold values, we present the averages of the transformed labels.
By adjusting these average values, we can explore the impact of both conservative and aggressive label transformations. 
Notably, setting the average of transformed labels to 1 equates to the DCMT model.

\begin{figure}[t]
    \centering
    \includegraphics[width=\columnwidth]{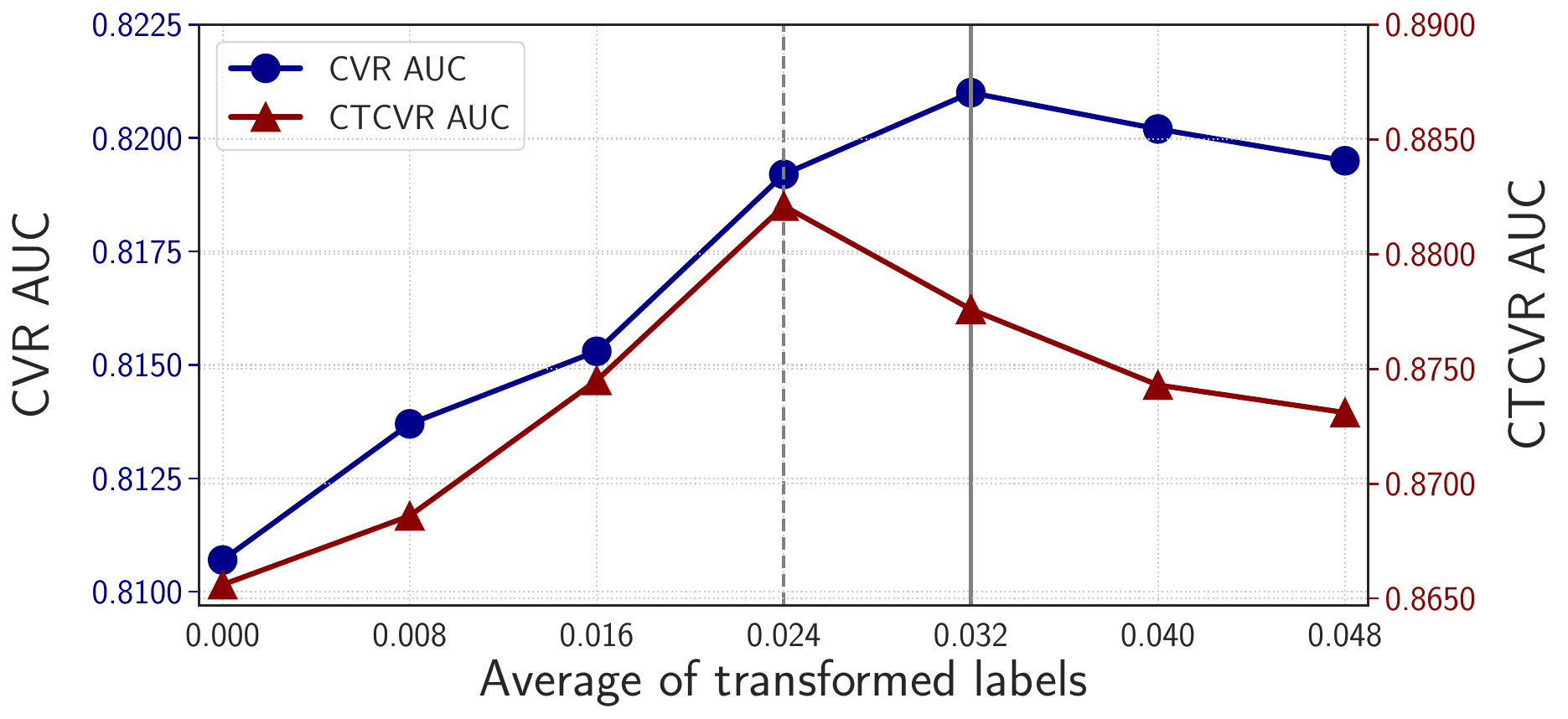}
    \caption{The CVR and CTCVR AUC for different thresholds on the AE-FR dataset.}
    \label{fig:rq3}
\end{figure}

We conduct experiments on the AE-FR dataset using an MLP with [512, 256, 128] layers as the backbone.
\cref{fig:rq3} shows the CVR and CTCVR AUC as functions of the average transformed label values.
The solid and dotted vertical lines indicate the averages of the max and ratio approaches, respectively.
The highest CVR and CTCVR AUCs are observed at average label values of 0.032 and 0.024, corresponding to the max and ratio approaches, respectively.
Our approaches result in optimal CVR and CTCVR AUC in the dataset, suggesting that our approaches provide sensible strategies for choosing the appropriate cut-off value.
Moreover, it implies that excessively or insufficiently transforming the labels for non-clicked samples to 1 can lead to a mismatch between the counterfactual and actual label distributions, ultimately resulting in a drop in performance.

\subsection{Ablation Study on Posterior Distribution (RQ4)}
\label{subsec:rq4}
We conduct ablation studies to investigate the effect of incorporating counterfactual inference on inferring pCVRs for non-clicked samples by implementing two variants of \ours{} and comparing their performance.
One model employs a naive imputation method, assuming that $Z$ does not exist.
Specifically, we pre-train a model using clicked samples with input features and click labels, then infer the pCVRs for the non-clicked samples. 
Afterward, the max approach for label transformation is applied to the values.
The other model adopts the proposed framework but skips the ``Abduction" step, which accurately estimates the posterior probability $p(Z \mid X, C, V)$. 
Instead, we assume $Z \sim \cN(0, I)$ and proceed with the steps starting from \cref{subsubsec:action}, followed by the max approach.

The experiments are conducted on the Ali-Express dataset, using CVR AUC as the evaluation metric, with results summarized in \cref{table:ablation_study}.
\ours{} significantly outperforms the variants across all datasets, where the improvement is particularly notable in the AE-FR and AE-US datasets. 
This demonstrates the effectiveness of our approach in accurately estimating the distribution of $Z$, leading to better model performance.
Additionally, we observe that naively assuming $Z \sim \cN(0, I)$ results in lower CVR AUC compared to not using $Z$ in the AE-ES and AE-FR datasets.
This suggests that such an assumption may degrade performance, emphasizing the importance of accurate estimations of posterior distribution.

\begin{table}[t]
     \setlength{\tabcolsep}{6pt}
    \centering
    \caption{Impact of $Z$ on the AE-ES dataset.  The best results are in boldface.
    }
    \label{table:ablation_study}
        \begin{tabular}{@{}rcccc@{}}
            \toprule
            \textbf{Method} & \textbf{AE-ES} & \textbf{AE-FR} & \textbf{AE-NL} & \textbf{AE-US} \\
            \midrule
            w/o $Z$ & 0.8284 &  0.8150 & 0.7893 & 0.8185 \\
            $Z \sim \cN(0, I)$ & 0.8269 &  0.8148 & 0.7904 & 0.8236 \\
            $Z \sim p(Z | X, C, V)$ & \textbf{0.8314} &  \textbf{0.8210} & \textbf{0.7951} & \textbf{0.8332} \\
            \bottomrule
        \end{tabular}%
\end{table}

\subsection{Performance on Latent Conversion Data (RQ5)}
\label{subsec:rq5}

\begin{figure*}[t]
    \centering
    \includegraphics[width=\linewidth]{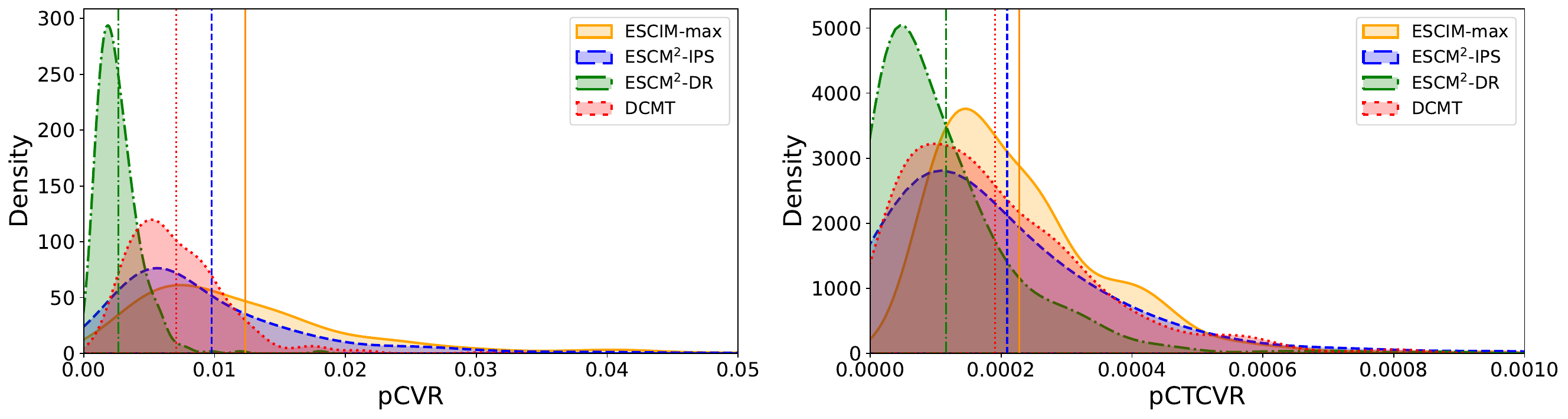}
    \caption{Distributions of pCVR and pCTCVR on latent conversion data in the Ali-CCP dataset}
    \label{fig:comp_ood_data}
\end{figure*}


\begin{table}[t]
    \centering
    \caption{Performance comparisons of the proposed model with baselines on latent conversion data in the Ali-CCP dataset. The backbone is an MLP with [512, 256, 128] layers. The best results are in boldface.
    }
    \begin{tabular}{@{}lcc@{}}
        \toprule
        \textbf{Method} & \textbf{CVR AUC} & \textbf{CTCVR AUC} \\
        \midrule 
            \textbf{ESCM$^2$-IPS} & 0.6724 & 0.6067 \\
            \textbf{ESCM$^2$-DR} & 0.6627 & 0.5983 \\
            \textbf{DCMT} & 0.6783 & 0.5926 \\
            \textbf{\ours-max} & \textbf{0.6905} & \textbf{0.6289} \\
        \bottomrule
    \end{tabular}
    \label{table:perf_unseen_data}
\end{table}

Training the model on counterfactual data, under the premise of ``Would the user have converted if he/she had clicked the recommended item?", primes it for more accurate predictions of user behaviors---especially for those who did not click during training and validation but exhibited click behavior during the test phase, which we define as \textit{latent conversion data}.
This data is instrumental in refining our algorithms to forecast latent conversion potential, ensuring our model can adapt to and detect these hidden behavioral shifts.
Therefore, we extract such data from the Ali-CCP dataset and measure CVR and CTCVR AUC for both baselines and \ours{}-max.
In addition, we plot distributions (using kernel density estimation) of pCVR and predicted CTCVR (pCTCVR) from the baselines and \ours{} for samples where both click and conversion labels are 1 and verify if the model returns higher prediction for such data.

The results are reported in \cref{table:perf_unseen_data} and \cref{fig:comp_ood_data}.
In \cref{fig:comp_ood_data}, the orange solid line depicts the distributions of pCVR and pCTCVR for \ours-max, while other lines represent the distributions for the baseline models: ESCM$^2$-IPS (blue dashed), ESCM$^2$-DR (green dash-dotted), and DCMT (red dotted).
The vertical lines indicate the mean of the pCVR and pCTCVR for the respective models.
We have the following observations:
\begin{itemize}[leftmargin=*]
    \item 
    \cref{table:perf_unseen_data} shows that our model outperforms the baselines in both CVR and CTCVR AUC on this dataset.
    Specifically, \ours-max achieves CVR and CTCVR AUC of 0.6905 and 0.6289, which are 1.80\% and 3.66\% higher than the best baseline performance, respectively.
    We confirm that ESCIM provides more accurate conversion predictions for users who did not click during the training and validation phases. 
    This demonstrates that \ours{}-max generalizes better to latent conversion samples compared to baselines and is more robust to sample selection bias.
    \item 
    \ours{}-max exhibits, in \cref{fig:comp_ood_data}, both CVR and CTCVR prediction distributions skewed more towards higher values compared to the baselines.
    This implies that \ours{} predicts higher (CT)CVRs for users who have converted in the test set, demonstrating better generalization performance for latent conversion users.
\end{itemize}

\section{Related Works}
\label{sec:realted_works}
\subsection{CVR Prediction}

Multi-task learning frameworks have been widely used to mitigate data sparsity and selection bias in CVR prediction. 
ESMM~\cite{Ma18Esmm} jointly optimizes CTR and CVR models, while later extensions~\cite{Wang22Escm2, Wen20Esm2} improve performance by incorporating unobserved samples or decomposing post-click behavior. 
However, these methods do not fully resolve selection bias in the non-clicked space.
Our method addresses this gap by accurately generating counterfactual conversion labels for non-clicked samples, improving label quality compared to DCMT~\cite{Zhu23Dcmt}, which naively assigns all labels as 1.
In parallel, multi-task models with advanced architectures~\cite{Ma18Mmoe, Tang20Ple} improve prediction by capturing task-specific patterns.
Nevertheless, these methods do not address the issue of MNAR.

\subsection{Causal Inference in Recommender Systems}
Recent studies have proposed incorporating causal inference into the domain of recommendation systems to analyze the genuine impact of specific components within systems.
Initial research mainly focused on correcting biases in implicit feedback, such as exposure bias~\cite{Schnabel16Ips}. 
Subsequently, many works have been proposed to address the issues of additional biases like amplification bias~\cite{Wang21DecRS} and popularity bias~\cite{Zhang21Pb} within training datasets using backdoor adjustment.
With the advancement of multi-task learning, new debiasing methods based on IPS or DR estimators have been introduced to improve the accuracy of CVR predictions~\cite{Zhang20Multi, Wang22Escm2}.
Moreover, recent works have employed the counterfactual inference to various recommendation domains, such as in top-$n$ recommendation \cite{Yang21topn} and out-of-distribution recommendation \cite{Wang22Cor}.

\section{Conclusion and Future Work}
\label{sec:conclusion}
We proposed a method for enhancing CVR prediction, called \ours{}, which adeptly addresses the inherent challenges of sample selection bias and data sparsity by generating and utilizing counterfactual conversion labels for non-clicked samples. 
Extensive experimentation on public datasets demonstrated the superiority of \ours{} over state-of-the-art methods.
The online A/B test further empirically validated its effectiveness of \ours{} in real-world scenarios. 
In addition, analysis of latent conversion data showcased the improved generalization performance of \ours{}.

For future work, there are two promising directions to explore. 
First, a more precise generation procedure for the counterfactual label could further enhance the accuracy of the CVR prediction. 
The process for inferring counterfactual CVR can be further refined, and more robust methods of converting counterfactual pCVR into hard labels can also be explored. 
Another approach focuses on mitigating the instability of the inverse propensity weighting, which corresponds to the reciprocal of CTR in this domain.
The effectiveness of \ours{} relies heavily on accurate CTR estimation, yet achieving this remains a significant challenge.

\section*{Acknowledgments}
This work was partly supported
by the IITP (RS-2022-II220953/25\%) and NRF (RS-2023-
00211904/50\%, RS-2023-00222663/25\%) grant funded by
the Korean government.

\bibliographystyle{IEEEtran}
\bibliography{sample-base}

\begin{thebibliography}{10}
\providecommand{\url}[1]{#1}
\csname url@samestyle\endcsname
\providecommand{\newblock}{\relax}
\providecommand{\bibinfo}[2]{#2}
\providecommand{\BIBentrySTDinterwordspacing}{\spaceskip=0pt\relax}
\providecommand{\BIBentryALTinterwordstretchfactor}{4}
\providecommand{\BIBentryALTinterwordspacing}{\spaceskip=\fontdimen2\font plus
\BIBentryALTinterwordstretchfactor\fontdimen3\font minus \fontdimen4\font\relax}
\providecommand{\BIBforeignlanguage}[2]{{%
\expandafter\ifx\csname l@#1\endcsname\relax
\typeout{** WARNING: IEEEtran.bst: No hyphenation pattern has been}%
\typeout{** loaded for the language `#1'. Using the pattern for}%
\typeout{** the default language instead.}%
\else
\language=\csname l@#1\endcsname
\fi
#2}}
\providecommand{\BIBdecl}{\relax}
\BIBdecl

\bibitem{Zhou19Dien}
G.~Zhou, N.~Mou, F.~Ying, Q.~Pi, W.~Bian, T.-C. Zhou, X.~Zhu, and K.~Gai, ``Deep interest evolution network for click-through rate prediction,'' in \emph{AAAI}.\hskip 1em plus 0.5em minus 0.4em\relax Honolulu, Hawaii, USA: MIT Press, 2019, pp. 5941--5948.

\bibitem{Ma18Esmm}
X.~Ma, L.~Zhao, G.~Huang, Z.~Wang, Z.~Hu, X.~Zhu, and K.~Gai, ``Entire space multi-task model: An effective approach for estimating post-click conversion rate,'' in \emph{SIGIR}.\hskip 1em plus 0.5em minus 0.4em\relax Ann Arbor, MI, USA: ACM, 2018, pp. 1137--–1140.

\bibitem{Wen20Esm2}
H.~Wen, J.~Zhang, Y.~Wang, F.~Lv, W.~Bao, Q.~Lin, and K.~Yang, ``Entire space multi-task modeling via post-click behavior decomposition for conversion rate prediction,'' in \emph{SIGIR}.\hskip 1em plus 0.5em minus 0.4em\relax Xi'an, China: ACM, 2020, pp. 2377--2386.

\bibitem{Pan18Fwfm}
J.~Pan, J.~Xu, A.~L. Ruiz, W.~Zhao, S.~Pan, Y.~Sun, and Q.~Lu, ``Field-weighted factorization machines for click-through rate prediction in display advertising,'' in \emph{WWW}.\hskip 1em plus 0.5em minus 0.4em\relax Lyon, France: ACM, 2018, pp. 1349--1357.

\bibitem{Wang22Escm2}
H.~Wang, T.-W. Chang, T.~Liu, J.~Huang, Z.~Chen, C.~Yu, R.~Li, and W.~Chu, ``{ESCM}$^2$: Entire space counterfactual multi-task model for post-click conversion rate estimation,'' in \emph{SIGIR}.\hskip 1em plus 0.5em minus 0.4em\relax Madrid, Spain: ACM, 2022, pp. 363--372.

\bibitem{Zhu23Dcmt}
F.~Zhu, M.~Zhong, X.~Yang, L.~Li, L.~Yu, T.~Zhang, J.~Zhou, C.~Chen, F.~Wu, G.~Liu, and Y.~Wang, ``{DCMT}: A direct entire-space causal multi-task framework for post-click conversion estimation,'' in \emph{ICDE}.\hskip 1em plus 0.5em minus 0.4em\relax San José, Costa Rica: ACM, 2023, pp. 3113--3125.

\bibitem{Pearl16Ci}
J.~Pearl, M.~Glymour, and N.~P. Jewell, \emph{Causal inference in statistics: A primer}.\hskip 1em plus 0.5em minus 0.4em\relax Chichester, UK: John Wiley \& Sons, 2016.

\bibitem{Correa21Nested}
\BIBentryALTinterwordspacing
J.~Correa, S.~Lee, and E.~Bareinboim, ``Nested counterfactual identification from arbitrary surrogate experiments,'' in \emph{NeurIPS}, M.~Ranzato, A.~Beygelzimer, Y.~Dauphin, P.~Liang, and J.~W. Vaughan, Eds., vol.~34.\hskip 1em plus 0.5em minus 0.4em\relax Curran Associates, Inc., 2021, pp. 6856--6867. [Online]. Available: \url{https://proceedings.neurips.cc/paper_files/paper/2021/file/36bedb6eb7152f39b16328448942822b-Paper.pdf}
\BIBentrySTDinterwordspacing

\bibitem{Yang21topn}
M.~Yang, Q.~Dai, Z.~Dong, X.~Chen, X.~He, and J.~Wang, ``Top-n recommendation with counterfactual user preference simulation,'' in \emph{CIKM}.\hskip 1em plus 0.5em minus 0.4em\relax Gold Coast, Queensland, Australia: ACM, 2021, pp. 2342--2351.

\bibitem{Wang22Cor}
W.~Wang, X.~Lin, F.~Feng, X.~He, M.~Lin, and T.-S. Chua, ``Causal representation learning for out-of-distribution recommendation,'' in \emph{WWW}.\hskip 1em plus 0.5em minus 0.4em\relax Virtual Event, Slovenia: ACM, 2021, pp. 3562--3571.

\bibitem{Blei17}
D.~M. Blei, A.~Kucukelbir, and J.~D. McAuliffe, ``Variational inference: A review for statisticians,'' \emph{Journal of the American Statistical Association}, vol. 112, pp. 859--877, 2017.

\bibitem{Kingma14}
D.~P. Kingma and M.~Welling, ``Auto-encoding variational bayes,'' in \emph{ICLR}.\hskip 1em plus 0.5em minus 0.4em\relax Banff, Canada: ICLR, 2014.

\bibitem{Liang20}
D.~Liang, R.~G. Krishnan, M.~D. Hoffman, and T.~Jebara, ``Variational autoencoders for collaborative filtering,'' in \emph{WWW}.\hskip 1em plus 0.5em minus 0.4em\relax Lyon, France: ACM, 2018, pp. 689--698.

\bibitem{Li20}
P.~Li, R.~Li, Q.~Da, A.-X. Zeng, and L.~Zhang, ``Improving multi-scenario learning to rank in e-commerce by exploiting task relationships in the label space,'' in \emph{CIKM}.\hskip 1em plus 0.5em minus 0.4em\relax Galway, Ireland: ACM, 2020, pp. 2605--2612.

\bibitem{Zhang20Multi}
W.~Zhang, W.~Bao, X.-Y. Liu, K.~Yang, Q.~Lin, H.~Wen, and R.~Ramezani, ``Large-scale causal approaches to debiasing post-click conversion rate estimation with multi-task learning,'' in \emph{WWW}.\hskip 1em plus 0.5em minus 0.4em\relax Virtual Event, Taiwan: ACM, 2020, pp. 2775--2781.

\bibitem{Guo17Deepfm}
H.~Guo, R.~Tang, Y.~Ye, Z.~Li, and X.~He, ``{DeepFM}: A factorization-machine based neural network for {CTR} prediction,'' in \emph{IJCAI}.\hskip 1em plus 0.5em minus 0.4em\relax Melbourne, Australia: International Joint Conferences on Artificial Intelligence Organization, 2017, pp. 1725--1731.

\bibitem{Song19Autoint}
W.~Song, C.~Shi, Z.~Xiao, Z.~Duan, Y.~Xu, M.~Zhang, and J.~Tang, ``{AutoInt}: Automatic feature interaction learning via self-attentive neural networks,'' in \emph{CIKM}.\hskip 1em plus 0.5em minus 0.4em\relax Beijing, China: ACM, 2019, pp. 1161--1170.

\bibitem{Wang21Dcnv2}
R.~Wang, R.~Shivanna, D.~Cheng, S.~Jain, D.~Lin, L.~Hong, and E.~Chi, ``{DCN V2}: Improved deep \& cross network and practical lessons for web-scale learning to rank systems,'' in \emph{WWW}.\hskip 1em plus 0.5em minus 0.4em\relax Virtual Event, Slovenia: ACM, 2021, pp. 1785--1797.

\bibitem{Ma18Mmoe}
J.~Ma, Z.~Zhao, X.~Yi, J.~Chen, L.~Hong, and E.~H. Chi, ``Modeling task relationships in multi-task learning with multi-gate mixture-of-experts,'' in \emph{KDD}.\hskip 1em plus 0.5em minus 0.4em\relax London, United Kingdom: ACM, 2018, pp. 1930--1939.

\bibitem{Tang20Ple}
H.~Tang, J.~Liu, M.~Zhao, and X.~Gong, ``Progressive layered extraction: A novel multi-task learning model for personalized recommendations,'' in \emph{RecSys}.\hskip 1em plus 0.5em minus 0.4em\relax Virtual Event: ACM, 2020, pp. 269--278.

\bibitem{Schnabel16Ips}
T.~Schnabel, A.~Swaminathan, A.~Singh, N.~Chandak, and T.~Joachims, ``Recommendations as treatments: Debiasing learning and evaluation,'' in \emph{ICML}.\hskip 1em plus 0.5em minus 0.4em\relax New York City, NY, USA: PMLR, 2016, pp. 1670--1679.

\bibitem{Wang21DecRS}
W.~Wang, F.~Feng, X.~He, X.~Wang, and T.-S. Chua, ``Deconfounded recommendation for alleviating bias amplification,'' in \emph{KDD}.\hskip 1em plus 0.5em minus 0.4em\relax Virtual Event, Singapore: ACM, 2021, pp. 1717--1725.

\bibitem{Zhang21Pb}
Y.~Zhang, F.~Feng, X.~He, T.~Wei, C.~Song, G.~Ling, and Y.~Zhang, ``Causal intervention for leveraging popularity bias in recommendation,'' in \emph{SIGIR}.\hskip 1em plus 0.5em minus 0.4em\relax Virtual Event, Canada: ACM, 2021, pp. 11--20.

\bibitem{KingBa15}
D.~Kingma and J.~Ba, ``Adam: A method for stochastic optimization,'' in \emph{ICLR}.\hskip 1em plus 0.5em minus 0.4em\relax San Diego, CA, USA: ICLR, 2015.

\end{thebibliography}

\section{Appendix}
\label{appendix}

\subsection{Experiments Details}
\label{appendix:exp_details}

\subsubsection{Description of Datasets}
\label{appendix:dataset_description}
\begin{table}[ht]
    \centering
    \caption{Statistics of datasets.}
    \label{table:datasets_stats}
    \begin{tabular}{c|c|c|c|c|c}
        \toprule
        \textbf{Dataset} & \textbf{\# Train} & \textbf{\# Val} & \textbf{\# Test} & \textbf{\# Click} & \textbf{\# Conv.} \\
        \midrule
        Industrial & 6.6 M & 0.7 M & 0.1 M & 0.7 M & 0.2 M \\
        Ali-CCP & 38.1M & 4.2M & 43.3M & 3.3M & 18.3K \\
        AE-ES & 21.1M & 2.2M & 9.3M & 0.8M & 19.0K \\
        AE-FR & 16.4M & 1.8M & 8.8M & 0.5M & 14.3K \\
        AE-NL & 11.0M & 1.2M & 5.6M & 0.4M & 13.8K \\
        AE-US & 18.0M & 2.0M & 7.5M & 0.5M & 10.9K \\
        \bottomrule
    \end{tabular}
\end{table}

The statistics of the datasets are described in \cref{table:datasets_stats}.
The Industrial dataset used in our study is derived from a 34-day record of our company's recommendation system, sequentially segmented into training, validation, and test datasets, comprising periods of 30 days, 3 days, and 1 day, respectively.
Due to the abundance of negative samples in the datasets, we apply downsampling to achieve a negative-to-positive sample click ratio of 5:1.
To secure data privacy, the dataset has been encrypted and anonymized, making it exclusively available for scholarly research.

\subsubsection{Implementation Details}
\label{appendix:imp_details}
For the MLP backbone, we adopted a simple three-layer structure with hidden units $[512, 256, 128]$ for the Ali-CCP dataset, $[64, 64, 32]$ for the Ali-Express dataset, employing Leaky ReLU activation.
For AutoInt, the number of layers and heads is 3 and 2, respectively, and the attention size is equivalent to the input size.
For DCN-V2, we used the parallel structure and two cross-layers. 
Adam \cite{KingBa15} is chosen as the optimizer, where the learning rate is $10^{-4}$ and the weight decay is $10^{-6}$.
L2 regularization is set to $10^{-4}$, and the dropout rate is 0.1. 
Hyperparameters were selected based on preliminary validation performance and fixed for all experiments, without additional tuning.
All experiments were conducted on a server equipped with 4 NVIDIA Tesla T4 GPUs and 256 GB of RAM, where training \ours{} on the Ali-CCP dataset took approximately 3 hours.

\begin{figure}[t]
    \centering
    \includegraphics[width=\columnwidth]{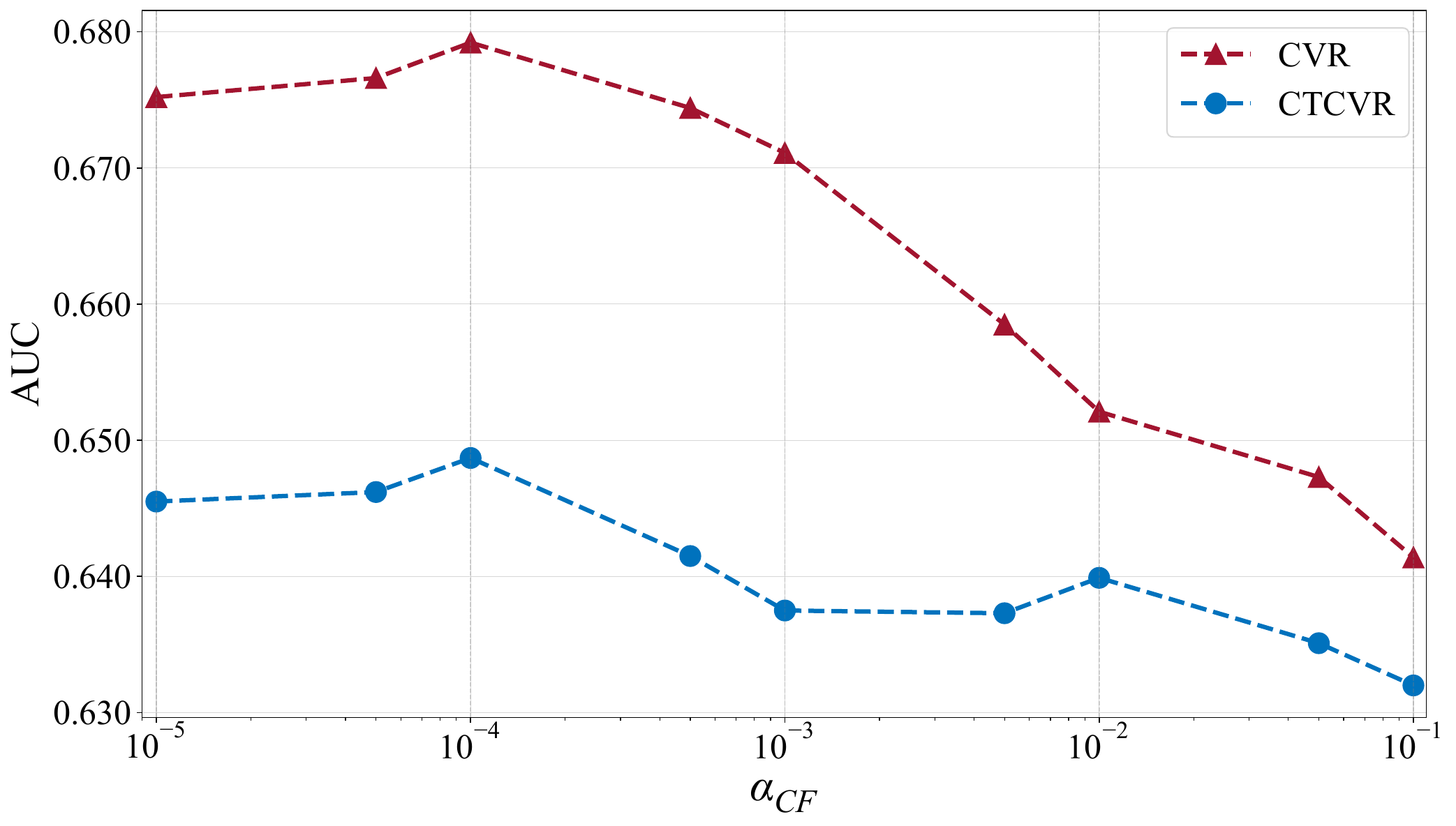}
    \caption{The CVR and CTCVR AUC for different values of $\alpha_{CF}$ on the Ali-CCP dataset.}
    \label{fig:aucs}
\end{figure}

\subsection{Parameter Sensitivity}
\label{subsec:param_sensitivity}

We explore the impact of $\alpha_{CF}$ on model performance by varying its value from $10^{-4}$ to $10^{-1}$ on a logarithmic scale, while fixing $\alpha_{F}$ to 0.1 for the Ali-CCP dataset.
The max approach is employed for the label transformation.
The results are depicted in \cref{fig:aucs}, where the red and blue line indicates the CVR and CTCVR AUC, respectively.
As illustrated in \cref{fig:aucs}, CVR AUC degrades noticeably as $\alpha_{CF}$ increases.
For example, in the Ali-CCP dataset, the CVR and CTCVR AUC are 0.6414 and 0.6320 when $\alpha_{CF}=10^{-1}$, which is 5.56\% and 2.57\% less than the best-performing AUC achieved when $\alpha_{CF}=10^{-4}$.
These results suggest that excessive reliance on counterfactual data can harm model performance, potentially due to noise or label inaccuracies in the generated samples.

\end{document}